# Ukrainian-To-English Folktale Corpus: Parallel Corpora Creation and Augmentation for Machine Translation in Low-Resource Languages

**Olena Burda-Lassen, Ph.D.**  oburdalassen@gmail.com
Independent Scholar, Colorado, United States

**Abstract**

Folktales are linguistically very rich and culturally significant in understanding the source language. Historically, only human translation has been used for translating folklore. Therefore, the number of translated texts is very sparse, which limits access to knowledge about cultural traditions and customs. We have created a new Ukrainian-To-English parallel corpus of familiar Ukrainian folktales based on available English translations and suggested several new ones. We offer a combined domain-specific approach to building and augmenting this corpus, considering the nature of the domain and differences in the purpose of human versus machine translation. Our corpus is word and sentence-aligned, allowing for the best curation of meaning, specifically tailored for use as training data for machine translation models.

## 1. Introduction

Machine translation has tremendous potential in connecting people and cultures. The Ukrainian language has an extensive collection of myths, legends, proverbs, songs, and folktales. They all represent the emotions, beliefs, and world views of Ukrainians.

In this paper, we focus on several widely known Ukrainian folktales, all of which are anonymous due to the nature of this genre. Furthermore, folktales are usually passed on from one generation to another, going back hundreds of years.

Interestingly, many available translations are rather transcreations, in which stories are retold and adapted to the target language and culture. We believe machine translation can be a useful supplemental tool in translating Ukrainian folklore, creating opportunities for more research and knowledge transfer about the Ukrainian language and culture. The first step in improving the machine translation performance of Ukrainian folktales is the creation of a high-quality corpus that addresses domain-specific nuances and challenges.

## 2. Parallel Corpus Creation and Augmentation

### 2.1. Available Resources Overview

Historically, Ukrainian has been considered a low-resource language with limited corpora resources (Grabar et al., 2018). However, the creation of WikiMatrix (Schwenk et al., 2021a) and CCMatrix (Schwenk et al., 2021b) has significantly improved access to training data for the Ukrainian language. Even when the size of the parallel corpus is significant, the smaller high-quality corpus can increase translation performance (Yıldız et al., 2014), especially within such a unique and challenging domain.

Therefore, we focused on carefully selecting familiar Ukrainian folktales available in English. One of the most extensive collections of English translations of Ukrainian fairytales and folktales is Project Gutenberg's *Cossack Fairy Tales and Folk Tales*[1]. This collection was initially published in 1894 and was edited and translated by Robert Nisbet Bain[2].

We were looking for available original Ukrainian texts containing culture-loaded words and word combinations (primarily from mythology and social life), which helped in the final decision about selecting respective English translations. In our future work, we will expand the available corpus to include more source and target texts.

Ukrainian original versions of selected folktales come from an online collection of stories for children[3] and blogs about Ukrainian traditions[4]. English translations are used from the Gutenberg Project and available translations of folktales[5].

We created this corpus to support a narrow domain of Ukrainian folklore. However, it could also work for the general translation of informational texts about Ukrainian culture. Recently, interest in the Ukrainian language and cultural knowledge has risen. We have found a website containing information about Ukrainian traditions, including many familiar folktales, legends, song lyrics, and stories about customs and holidays. This website is written in Ukrainian and is recommended for English speakers to read in its machine-translated version[6]. It is an excellent example of a practical machine translation application.

Unfortunately, only a limited number of folktales are translated into English. Machine translation models trained using the proposed corpus could fill this gap and help spread knowledge about the Ukrainian language and culture.

**2.2. Methods**

Our corpus consists of Ukrainian and English versions of 4 popular folktales: "The Mitten," "The Straw Ox," "The Bully Goat," and "Oh: The Tsar of the Forest."

The total number of aligned pairs of sentences and words is 400: the number of English words is 6,800, and the number of Ukrainian words is 4,157. This corpus is the start of our new project, and the number of parallel texts will be increasing consistently.

We have reviewed several available English versions of a well-known folktale, *The Mitten*[7], retold by Jan Brett, as well as *The Mitten: An Old Ukrainian Folktale*[8], by Alvin Tresselt, and decided to include our own, more literal version of the translation of this folktale.

While available English translations are poetic and commonly accepted in the target language space, we have proposed a more semantic translation instead of its adaptation. To be used as training data for machine translation models, source and target sentences must be translated as accurately as possible.

Due to the nature of this research, we needed to do a substantial amount of manual work related to curating training data.

---

[1] https://www.gutenberg.org/cache/epub/29672/pg29672.txt
[2] https://publicdomainreview.org/collection/cossack-fairy-tales-1916
[3] https://kazky.org.ua/
[4] https://carterhaughschool.com/the-fairy-tales-of-ukraine/
[5] https://pdfslide.net/documents/a-ukrainian-folk-tale-the-bully-goat-.html
[6] https://traditions-in-ua.translate.goog/?_x_tr_sl=uk&_x_tr_tl=en&_x_tr_hl=en&_x_tr_pto=sc
[7] https://janbrett.com/bookstores/mitten_book.htm
[8] Alvin Tresselt. 1964. The mitten: an old Ukrainian folktale. New York: Lothrop, Lee & Shepard Co., Inc.

After manually selecting each sentence in the source language and reviewing the equivalent in the target language, we have compiled sentence pairs. In some cases, source sentences and target language translations contained more or fewer sentences and were out of order or lossy. Therefore, we have aligned them according to the source language (for example, if one Ukrainian sentence was translated into two English sentences, we aligned them by the Ukrainian sentence). The English language column appears first in the corpus for easier corpus access and review by English speakers.

We have also aligned corpus by words to finetune the domain knowledge transfer. The examples we have selected were extracted from the corpus sentences and are culture-loaded words and word combinations, describing food, mythological creatures, and animals, for example, *мед-вино* ("med-vyno": "beer and mead"), *Мавка* ("Mavka": "Mavka, the forest spirit"), and *вовчик-братик* ("vovchyk-bratyk": "brother-wolf").

### 2.3. Findings

We believe that the success and accuracy of the machine translation system depend on the high accuracy of the rarely used source words. While most common phrases are already being translated accurately by available machine translation engines, it is the rare or cultural terms that get missed or misinterpreted by these engines. Adding an extra layer of culturally significant information can only improve the outcome of the translation process.

In the folktale "The Mitten," there are several proper names of animals consisting of their names and characteristic behavior traits, for example, *Мишка-шкряботушка* ("Myshka-shkryabotushka"). Therefore, we have proposed the translation "Scratching Mouse." The literal meaning of it is "the mouse that scratches on things." Hence, the term "Scratching Mouse," in our opinion, is semantically and stylistically more fitting for machine translation models.

A similar example of another hyphenated compound word from the story "The Mitten" is *Ведмідь-набрідь* ("Vedmid-nabrid"). Again, we suggest translating it to "Bear, the Wanderer." Both of these examples use loan translation with an element of expansion, which serves the informative purpose of corpus creation, tailored explicitly for machine translation systems.

In our corpus, we have also included translations of the words mentioned above by another translator Iryna Zheleznova[9]: "Crunch-Munch the Mouse" and "Grumbly-Rumbly the Bear." These terms work well for the English translation of this folktale in children's literature. It is rhymed and catchy, reflecting the target text's desired presentation.

Another example of an aligned word pair illustrates various translation methods of culture-loaded terms: *Мавка* ("Mavka"), one of the most widely known spirits in Ukrainian mythology. Mavka is a female forest spirit.

We have encountered the following translations of this mythology term: "Mavka" (transliteration), "water-nixie" and "nixie" (adaptation and generalization). Therefore, we propose using transliteration plus expansion to incorporate essential knowledge about this mythological creature: "Mavka, the forest spirit."

We have replaced archaic personal pronouns *thou, thee, thy, thine,* and *ye,* found in the English translation, with equivalent modern English pronouns. The source text does not include archaic pronouns, so we decided to omit their use in the corpus.

We hope that applying these augmentation techniques will further increase the quality of this parallel corpus. Furthermore, numerous examples of domain-specific translations can help train the machine translation model and increase accuracy, especially since examples are carefully curated and hand-picked.

---

[9] https://storytellingforeveryone.net/tag/cumulative-story/

## 3. Conclusions

Further research is necessary to create more extensive corpora, which we plan to conduct since only a very sparse number of corpora is available in the Ukrainian folktale domain.

However, contrary to the human translation methodology of folklore, machine translation techniques are more literal and descriptive.

We have aligned language pairs by sentences and words during parallel corpus creation to increase training data accuracy. We have observed a need for a significant difference between human and machine translation techniques within the folktale domain.

Ukrainian-To-English Folktale Corpus is publicly available online[10]. We also plan on researching the performance of this corpus on several machine translation models in the future.

## References

Genzel Dmitriy, Jakob Uszkoreit, and Franz Och. 2010. "Poetic" Statistical Machine Translation: Rhyme and Meter. In *Proceedings of the 2010 Conference on Empirical Methods in Natural Language Processing*, pages 158–166, Cambridge, MA. Association for Computational Linguistics.

Grabar Natalia, Olga Kanishcheva, Thierry Hamon. Multilingual aligned corpus with Ukrainian as the target language. *SLAVICORP*, Sep 2018, Prague, Czech Republic. ffhalshs-01968343f

Sánchez-Cartagena Víctor M., Miquel Esplà-Gomis, Juan Antonio Pérez-Ortiz, and Felipe Sánchez-Martínez. 2021. Rethinking Data Augmentation for Low-Resource Neural Machine Translation: A Multi-Task Learning Approach. In *Proceedings of the 2021 Conference on Empirical Methods in Natural Language Processing*, pages 8502–8516, Online and Punta Cana, Dominican Republic. Association for Computational Linguistics.

Schwenk Holger, Vishrav Chaudhary, Shuo Sun, Hongyu Gong, and Francisco Guzmán. 2021. WikiMatrix: Mining 135M Parallel Sentences in 1620 Language Pairs from Wikipedia. In *Proceedings of the 16th Conference of the European Chapter of the Association for Computational Linguistics:* Main Volume, pages 1351–1361, Online. Association for Computational Linguistics.

Schwenk Holger, Guillaume Wenzek, Sergey Edunov, Edouard Grave, Armand Joulin, and Angela Fan. 2021. CCMatrix: Mining Billions of High-Quality Parallel Sentences on the Web. In *Proceedings of the 59th Annual Meeting of the Association for Computational Linguistics and the 11th International Joint Conference on Natural Language Processing* (Volume 1: Long Papers), pages 6490–6500, Online. Association for Computational Linguistics.

Yıldız, Eray & Tantuğ, Ahmet & Diri, Banu. (2014). The Effect of Parallel Corpus Quality vs Size in English-To-Turkish SMT. *Computer Science & Information Technology*. 4. 21-30. 10.5121/csit.2014.4710.

---

[10] https://github.com/Ukrainian-To-English-Corpora/Folktale_corpus